\documentclass[10pt,twocolumn,letterpaper]{article}

\usepackage{wacv}
\usepackage{times}
\usepackage{epsfig}
\usepackage{graphicx}
\usepackage{amsmath}
\usepackage{amssymb}
\usepackage{booktabs}
\def\wacvPaperID{392} % Enter the WACV Paper ID here

\wacvalgorithmstrack   % Uncomment this line if you are submitting to the Algorithms Track.
\wacvfinalcopy % *** Uncomment this line for the final submission

\ifwacvfinal
\usepackage[breaklinks=true,bookmarks=false]{hyperref}
\else
\usepackage[pagebackref=true,breaklinks=true,colorlinks,bookmarks=false]{hyperref}
\fi

\begin{document}

%%%%%%%%% TITLE
\title{Style-Guided Inference of Transformer for High-resolution Image Synthesis}

\author{Jonghwa Yim\qquad\qquad\qquad\qquad Minjae Kim\\
Vision AI Lab., AI Center, NCSOFT\\
South Korea\\
{\tt\small \{jonghwayim, minjaekim\}@ncsoft.com}
%\and
%Minjae Kim\\
%NCSOFT AI Center\\
%{\tt\small minjaekim@ncsoft.com}
}

\maketitle
\thispagestyle{empty}

%%%%%%%%% ABSTRACT
\begin{abstract}
Transformer is eminently suitable for auto-regressive image synthesis which predicts discrete value from the past values recursively to make up full image.
Especially, combined with vector quantised latent representation, the state-of-the-art auto-regressive transformer displays realistic high-resolution images. 
However, sampling the latent code from discrete probability distribution makes the output unpredictable. Therefore, it requires to generate lots of diverse samples to acquire desired outputs. 
To alleviate the process of generating lots of samples repetitively, in this article, we propose to take a desired output, a style image, as an additional condition without re-training the transformer. To this end, our method transfers the style to a probability constraint to re-balance the prior, thereby specifying the target distribution instead of the original prior. Thus, generated samples from the re-balanced prior have similar styles to reference style.
In practice, we can choose either an image or a category of images as an additional condition. In our qualitative assessment, we show that styles of majority of outputs are similar to the input style.
\end{abstract}

%%%%%%%%% BODY TEXT

%===========================================================
\section{Introduction}

Image generation is a task that learns a mapping from a prior density to real image distribution. Recent works of deep generative models developed for this task derive from two major branches; one is Generative Adversarial Nets (GAN)~\cite{goodfellow2014generative} which is a pioneering research of implicit density models; the other is explicit density models which include Variational Auto-Encoder (VAE)~\cite{kingma2013auto} and Autoregressive generative models. 
Until recently, studies derive from GAN and VAE show distinguished achievement in image quality, however, they yet have a couple of limitations; first, they have a difficulty achieving high resolution and satisfactory output quality due to inherent limitations in either loss design or structurally restricted receptive field; second, the training process is relatively unstable.

To improve the problems of GAN and VAE based methods, another direction of research heads towards more direct formulation. To formulate tractable density, PixelRNN~\cite{van2016pixel} proposes an auto-regressive architecture to compute the global relationship explicitly and model the image components recursively. Thanks to the advantage in viewing global context, this research stably produces locally and globally coherent results. Following research, Conditional PixelCNN~\cite{oord2016conditional}, introduces a condition in representing the target distribution and displays qualitatively coherent results to given labels.
Despite solving the two major problems in GAN and VAE, its computation complexity is proportional to the square of the output size, hampering extending the size of an architecture and output resolution.

Recently, a transformer architecture from previous research~\cite{vaswani2017attention} provides clue for the above scalability problem in a tractable density model. By replacing recursive computation for long range relationships to transformer with masked self-attention module, the study achieves both of full receptive field and computational efficiency. Thus, following researches start adopting transformer instead of the recursive architecture. The research~\cite{parmar2018image} introduces transformer in image generation task and shows comparable or even better performance than recursive neural network (RNN) based architectures with a better computational efficiency. However, the limitation of these methods lies in insufficient quality compared to GAN-based methods since modeling low-level vision, pixel values in most cases, is qualitatively inaccurate compared to GAN-based methods.

Instead of predicting low-level vision, predicting quantised latent representation becomes a substitute. The first of this appears in VAE approaches. The previous researches~\cite{van2017neural}, \cite{razavi2019generating} introduce vector quantisation to latent representation in VAE, called Vector Quantised-Variational AutoEncoder (VQ-VAE), thereby the latent prior becomes discrete and categorical. Inspired by this study, the recent study~\cite{esser2021taming} adopts VQ-VAE to represent latent representation while the transformer infers quantized code instead of pixel value. Hence, the output of the transformer is a prior, probability density over discrete codebook. After the transformer infers the prior, sampling a code from the prior probability creates a sequence of codes which is deterministically translated to an image.
Because a quantised latent representation relieves the transformer from modeling the complex low-level vision, instead it can focus more on learning the global context. With the advantage of efficient role allocation, this study displays high-resolution and natural-looking outputs. However, what would be generated from prior sampling is unknown until it generates and decodes the full sequence of quantised codes, which is much more time consuming as the target resolution is getting higher. 

To resolve the unpredictable behavior of sampling from prior density, our proposal confers controllability on the above research. We propose a novel method to guide prior sampling by adopting a style image as a conditional input. To this end, we reformulate conditional generative model to posterior expression where conditional input is a given observation. Therefore, our method guides the inference of pre-trained transformer using a conditional input without re-training the model. 
Our method inherits the merits of the previous transformer-based generative approach, creating high-resolution photo-realistic output images. Besides, our method accepts conditional input image and generates outputs of similar styles to the given condition.

Our method can work both on a non-conditional and conditional pre-trained transformer, and its output is visually well persuasive, especially in landscape scenes where we can feed a semantic label map as a condition.
In the evaluation section, we show experiments of a single image and multiple images as input styles. In the case of multiple images, we perform sampling and averaging the distribution of latent features over the entire images.

%===========================================================
\section{Related Works}

{\bf GAN-based Image Synthesis.}
Over the last years, the field of image synthesis has been massively researched. Since the debut of Generative Adversarial Network~\cite{goodfellow2014generative}, this field has progressed to higher resolution and better quality than ever before. However, the original GAN holds a few drawbacks; the training process is unstable, enlarging the resolution is limited by the architectural design and machine to run the model, and the output quality is not yet that of the real world. Recently, a few cornerstone researches break the limitations and enable to synthesize the real-world scene.
The pioneering research, PG-GAN~\cite{karras2018progressive}, suggests GAN generating higher resolution by progressively stacking more layers and consequently enlarging the resolution. 
Meanwhile, to alleviate the instability in training process, we can find various former researches including Wasserstain GAN~\cite{arjovsky2017wasserstein}, PGGAN~\cite{karras2018progressive}, SN-GAN~\cite{miyato2018spectral}, techniques for training GANs~\cite{salimans2016improved}, LSGAN~\cite{mao2017least}, and BEGAN~\cite{berthelot2017began}. 
Improving the architecture is another branch of research; DCGAN~\cite{radford2015unsupervised} embraces convolutional architecture, SA-GAN~\cite{zhang2019self} adds self-attention blocks to its backbone, and BigGAN~\cite{brock2018large} composes many previously-existing modules such as self-attention, spectral normalization, and a discriminator with projection~\cite{miyato2018cgans}. 
Lastly, StyleGAN~\cite{karras2019style} is a foundation in its way. It proposes disentangled latent priors from mapping network instead of random noise for attribute control, stochastic variation, and better output quality.

{\bf Conditional Generative Model.} 
Conditional image generation refers to the task that generates images conditioning on a particular input data. For example, conditioning on a class label \cite{mirza2014conditional}, \cite{odena2017conditional}, \cite{brock2018large}, \cite{mescheder2018training}, \cite{miyato2018cgans} allows the generated samples to be in the class. Another case of conditional generative models \cite{reed2016generative}, \cite{zhang2017stackgan}, \cite{hong2018inferring}, \cite{xu2018attngan} is based on text to synthesize images. Recently, many researches \cite{wang2018high}, \cite{liu2019learning}, \cite{park2019semantic}, \cite{dundar2020panoptic}, \cite{tang2020dual}, \cite{zheng2020example}, \cite{zhu2020sean}, \cite{zhu2020semantically}, \cite{tan2021efficient}, \cite{tan2021diverse} use semantic segmentation map to indicate what to draw. SPADE~\cite{park2019semantic} suggests a spaital-varying normalization layer and becomes a significant backbone for the following researches. 
CC-FPSE~\cite{liu2019learning} applies spatially-varying convolutional kernels which are based on the semantic label map to fully utilize semantic layout information to generate high-quality and semantically high-fidelity images. SEAN~\cite{zhu2020sean} displays per-semantic styling by applying region-adaptive normalization. CLADE~\cite{tan2021efficient} proposes class-adaptive normalization layer. INADE~\cite{tan2021diverse} presents an instance-adaptive modulation sampling and improves diversity. 
Our work also extends the series of conditional generative model, except that it suggests guided inference to follow a given condition instead of architecturally accepting conditional input with additional parameters and requiring full training process.

\begin{figure*}[!ht]
\centering
\includegraphics[width=0.8\textwidth]{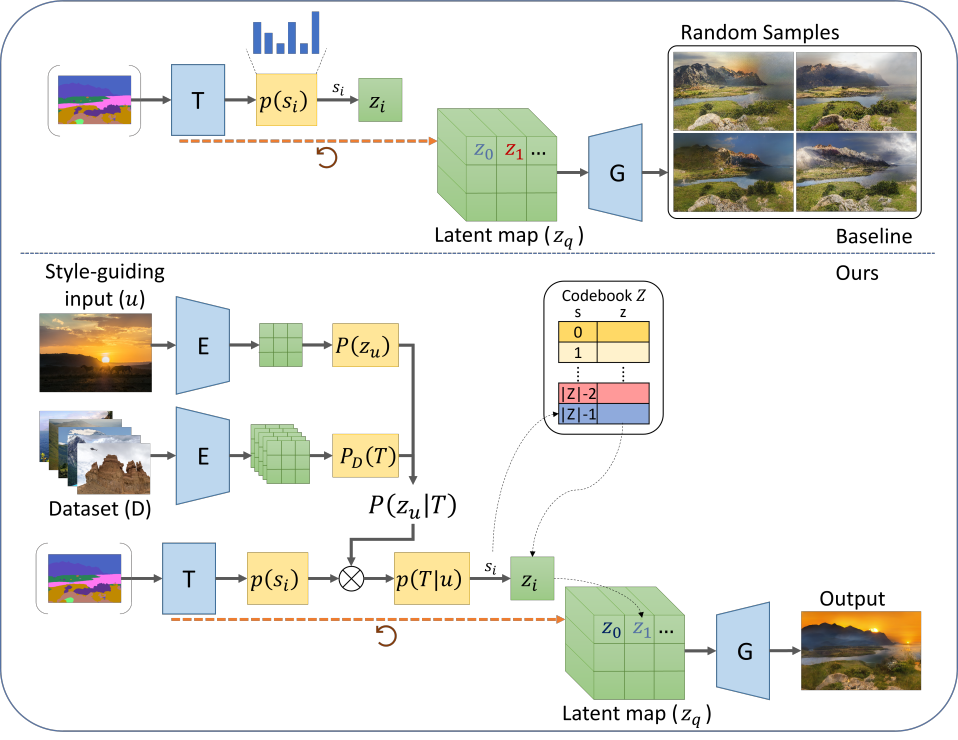}
\caption{{\bf System overview of image-generative process of baseline method~\cite{esser2021taming} (top) and our method (bottom).} The major difference from the baseline work is applying style likelihood $P(z_u|T)$ in the middle of the process to apply an additional condition $u$. $P(z_u)$ and $P_{D}(T)$ are probability distributions of all indices in latent maps. Latent map for $P(z_u)$ is from conditional image $u$. Latent maps for $P_{D}(T)$ are from images in training dataset $D$.
Encoder $E$, decoder $G$, and transformer $T$ are pre-trained.
In case of landscape synthesis, baseline method accepts semantic segmentation map as a conditional input, while in case of face synthesis, baseline is unconditional.}
\label{fig:fig1}
\end{figure*}

{\bf Autoregressive Generative Model.} 
Autoregressive model formulates a tractable density function as a generative approach. Due to its simplicity and directness, it is broadly used in image and audio generation. In image generation, autoregressive deep generative model predicts the probability over discrete pixel value regressively using the architectures called PixelRNN and PixelCNN \cite{van2016pixel}, \cite{oord2016conditional}, \cite{chen2018pixelsnail}. In audio generation, WaveNet~\cite{oord2016wavenet} introduces PixelCNN architecture to the 1D audio data by applying dilated causal convolution layer, followed by many researches~\cite{wang2017tacotron}, \cite{arik2017deep}, \cite{shen2018natural}, \cite{oord2018parallel} that improve the quality of synthesis. Afterwards, a transformer architecture~\cite{vaswani2017attention} is newly introduced in autoregressive generative method~\cite{parmar2018image} replacing the former RNN based architecture. Transformer becomes prevalent and also appears in GAN based approaches \cite{zhang2019self}, \cite{hudson2021ganformer}. Recently, the state-of-the-art study, Taming Transformer~\cite{esser2021taming}, adopts vector quantisation introduced in VQ-VAE~\cite{van2017neural}, \cite{razavi2019generating} and predicts probability density over the entries of a quantised latent vector instead of direct prediction of a discrete pixel value. Thanks to lessen the burden of transformer from predicting low-level complex vision feature, this work generates photo-realistic and high-resolution scenes. 

{\bf Arbitrary Photorealistic Style Transfer.} Arbitrary photorealistic style transfer is partially in line with our method in that it stylizes the output, though it requires a real input image.
Some recent studies~\cite{luan2017deep}, \cite{li2017universal}, \cite{li2018closed}, \cite{sheng2018avatar}, \cite{yoo2019photorealistic}, \cite{yim2020filter}, \cite{park2020swapping}, \cite{cheng2021style} introduce various methods to transfer style from a single image, but they demand a large dataset for training. 
Besides, they often yield undesirable transfer of color and texture due to the nature of the neural network.

%===========================================================
\section{Method}

%------------------------------------------------------------------------- 
\subsection{Baseline Work}

The upper half of Figure~\ref{fig:fig1} overviews the baseline method~\cite{esser2021taming}. 
The baseline model is possible to be both unconditional and conditional. Specifically, it is conditional upon a semantic label map in the case of landscape synthesis. 
The training process of the baseline is a two-stage. In the first stage, encoder $E$ and decoder $G$ are trained. While in the second stage, transformer $T$ is trained, and $E$ is omitted later.
$Z$ is a discrete codebook of size $|Z|$.
Transformer model $T$ auto-regressively outputs a probability distribution $p(s_i)$ over indices $s$ in $Z$, where an index $s_i$ is random-sampled. 
Latent code $z_i$ is a paired entry of $s_i$ in $Z$. $z_i$ fills the $i$-th entry in the latent map $z_q$.
As a result of an auto-regressive process, $z_q$ is formed and fed into $G$ to synthesize the output. 
In this process, random sampling from $p(s_i)$ makes the output unpredictable. Thus it is hard to find desired styles such as $u$ from random samples.

%------------------------------------------------------------------------- 
\subsection{Style-guided Inference of Transformer}

Our method aims for high-resolution conditionally guided image generation without re-training the transformer. On top of the latest progress of transformer~\cite{esser2021taming}, which incorporates an auto-regressive prediction and a quantized prior, our method will be a missing-piece to guide the inference of transformer conditionally and to stylize the output. 
To this end, we present a method to control the probability distribution over an auto-regressive latent prediction. 

The bottom half of Figure~\ref{fig:fig1} indicates an overview of the proposed method. 
$P(z_u)$ is a probability distribution of indices in the latent map of style image $u$, and $P_{D}(T)$ is the model's probability distribution of all indices in latent maps of images sampled from the training dataset $D$. 
A style likelihood $P(z_u|T)$ is drawn from $P(z_u)$ and $P_{D}(T)$. Then, it is repetitively multiplied to $p(s_i)$ to form a posterior $P(T|z_u)$. (Throughout the paper, we write $p(s_i)$ as $P(T)$ to denote a model's prior probability distribution.) 
From $P(T|z_u)$, a style-guided index $s_i$ is sampled, and $z^{(i)}$ is the paired entry of $s_i$. 

To be specific, if we condition $P(T)$, the inference of pre-trained model, on a style input $u$, then conditional probability is written as $P(T|u)$.
Previously, to make $T$ represent this distribution, $T$ needed supervised learning on a style-labeled dataset. Our method, however, proposes the conditionally guided-inference of pre-trained transformer without learning such condition. Without re-train, we represent the given problem to a style likelihood and a model's prior:
(Note that $\odot$ and $\oslash$ we use in this paper are element-wise multiplication and division of two vectors; $i$-th entries of $A \odot B$ and $A \oslash B$ are $A_i \times B_i$ and $A_i / B_i$ each)
\begin{align}
  P(T|u) \propto P(u|T) P(T) \text{.}
\label{eq:eq1_reformulation}
\end{align}
Instead of $u$, we replace it with the probability distribution of latent codes, $P(z_u)$, by ignoring the sequence of latent map $z_u$ and expressing all indices in $z_u$ as a probability distribution. (Note that the generator part is deterministic, and the specific sequence of $z_u$ restores the correct $u$.) However, ignoring the sequence makes $P(z_u)$ ill-posed to rebuild $u$. Instead, we rely on the last term, $P(T)$, to build the sequence as this term is recursive. 
In the same way we get $P(z_u)$, we also obtain the probability distribution of $T$ of the training dataset $D$, $P_D(T)$, by counting indices in latent maps from images in $D$. 
Then, we can view the likelihood term is proportional to the events of latent indices of conditional input $u$ divided by model's general event counts: $P(z_u) \oslash P_D(T)$. 
Accordingly, the problem is rewritten into 
\begin{align}
  P(z_u|T) \odot P(T) \propto \{P(z_u) \oslash P_D(T)\} \odot P(T) \text{,}
\label{eq:eq2_reformulation2}
\end{align}
where $P(\cdot) \in \mathcal{R}^{|Z|}$, probability over the codebook indices.

Practically, instead of calculating all latent maps from images in $D$ to get $P_D(T)$, we perform Monte-Carlo sampling and take the average, $1/K\sum_{k=1}^{K}P(z_{x_k})$, for all $K$ number of randomly sampled $x$ from the training dataset $D$.
Hence, the Eq.~\eqref{eq:eq2_reformulation2} becomes
\begin{align}
  P(z_u|T)P(T) 
  \propto  P(z_u) \oslash \{ \frac{1}{K}\sum_{k=1}^{K}P(z_{x_k}) \} \odot P(T)
\end{align}
where $\sum$ is element-wise summation.

Consequently, as the auto-regressive generative process continues, $P(z)$, the distribution of sampled indices from $P(T|u)$, is getting closer to $P(z_u)$, making the style of the generated sample similar to $u$.

In practice, to reinforce accuracy lost from dropping the spatial and sequential information when representing $P(u)$ as $P(z_u)$, we can use a semantic label map or coordinates in latent map. 
For example, when we have a semantic label map as a conditional input, we use it to count $P(z_u|T)$ partially by semantic regions: $P(z_{u,j})\oslash P_D(T_j)$ for $j$ labeled semantic region. (So we have $J$ number of style likelihoods where $J$ is the number of semantic labels.) In this way, we can accurately transfer the style of each semantic to a corresponding semantic.
In the following chapter, we show experiments on single style image and a categorized set of style images. In case of a categorized set of style images, we can compute the average style likelihood.

%===========================================================
\section{Experiments}

In this section, firstly, we design experiments on a landscape images because these have visually distinguishable styles like snowy or rocky mountain.
To prepare a landscape dataset for training, we collect 80k images from various sources, including Flickr, Pixabay, and Google search. Then, we generate pseudo-semantic labels using the well-known semantic segmentation method, DeepLabv3~\cite{chen2017rethinking} with ResNeSt-101~\cite{zhang2022resnest} backbone. 
As a base architecture for image generation, we configure an autoregressive transformer conditioning on a semantic label map presented by the previous paper~\cite{esser2021taming}. To get the model's probability $P_D(T)$ over the training dataset, we randomly select $K=700$ number of images from dataset.
Since our method can accept one or more style inputs, firstly, we show experiments on a single style image, then on a style-categorized set of multiple images sequentially for diversity. Note that we can also choose different source of style for each semantic.

\begin{figure*}
\centering
\includegraphics[width=\textwidth]{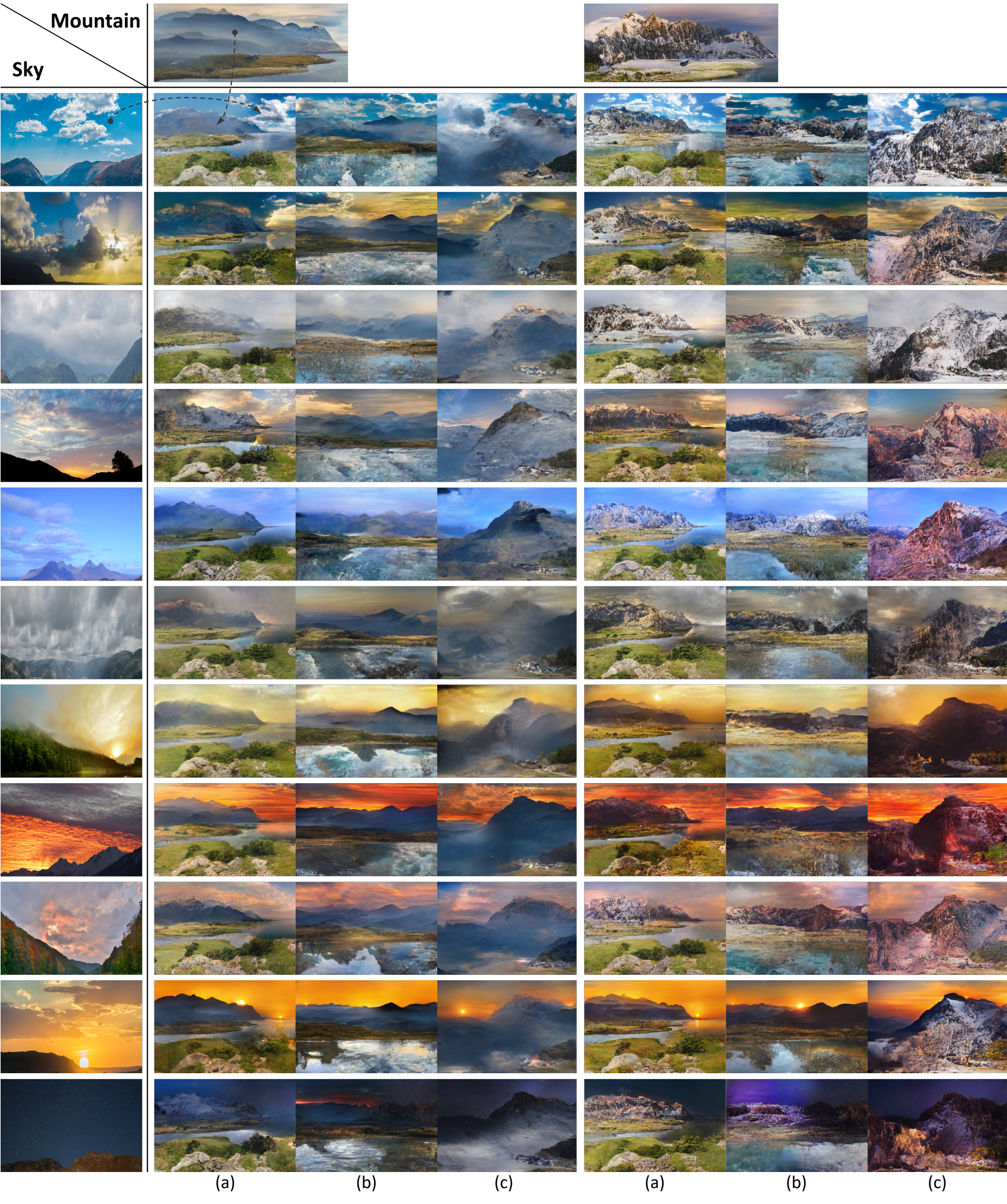}
\caption{{\bf Qualitative results of our proposed method.} Images on the left are style references of the sky, and on the top are styles of the mountain. Style of one of the sky references is applied to where semantic label matches the sky, while one of the mountain references is applied to where the semantic label matches the mountain. Three different semantic types (a), (b), and (c) are used. Thus, each cell is a mixture of semantically one of (a), (b), and (c), sky from sky reference on the left, and mountain from mountain reference on the top.}
\label{fig:fig3}
\end{figure*}

In the first experiment shown in Figure~\ref{fig:fig3}, we experiment with our method conditioned on a single style image. 
For exact modeling of style likelihood probability distribution from a style image, we divide regions by semantic labels and extract the style likelihood regionally. Then, we apply each distribution of semantic labels to the target semantic label, which we call semantic-aware stylization. 
For this experiment, we choose 11 different styles for the sky from the Flickr dataset and two different styles for the mountain from randomly synthesized samples. Then, style from one of the sky images and one of the mountain images are applied to the sky and mountain semantics each. 
Thus, the results have stylized sky and mountain in each image. 
(It is also possible to stylize more semantic labels with different style images.)
As a conditional input, we prepare three types of semantic label maps, (a), (b), and (c). (It can be inferred from result images.)
We repeat the guided-synthesis four times and pick the best one for each result image.

\begin{figure}
\centering
\includegraphics[width=\linewidth]{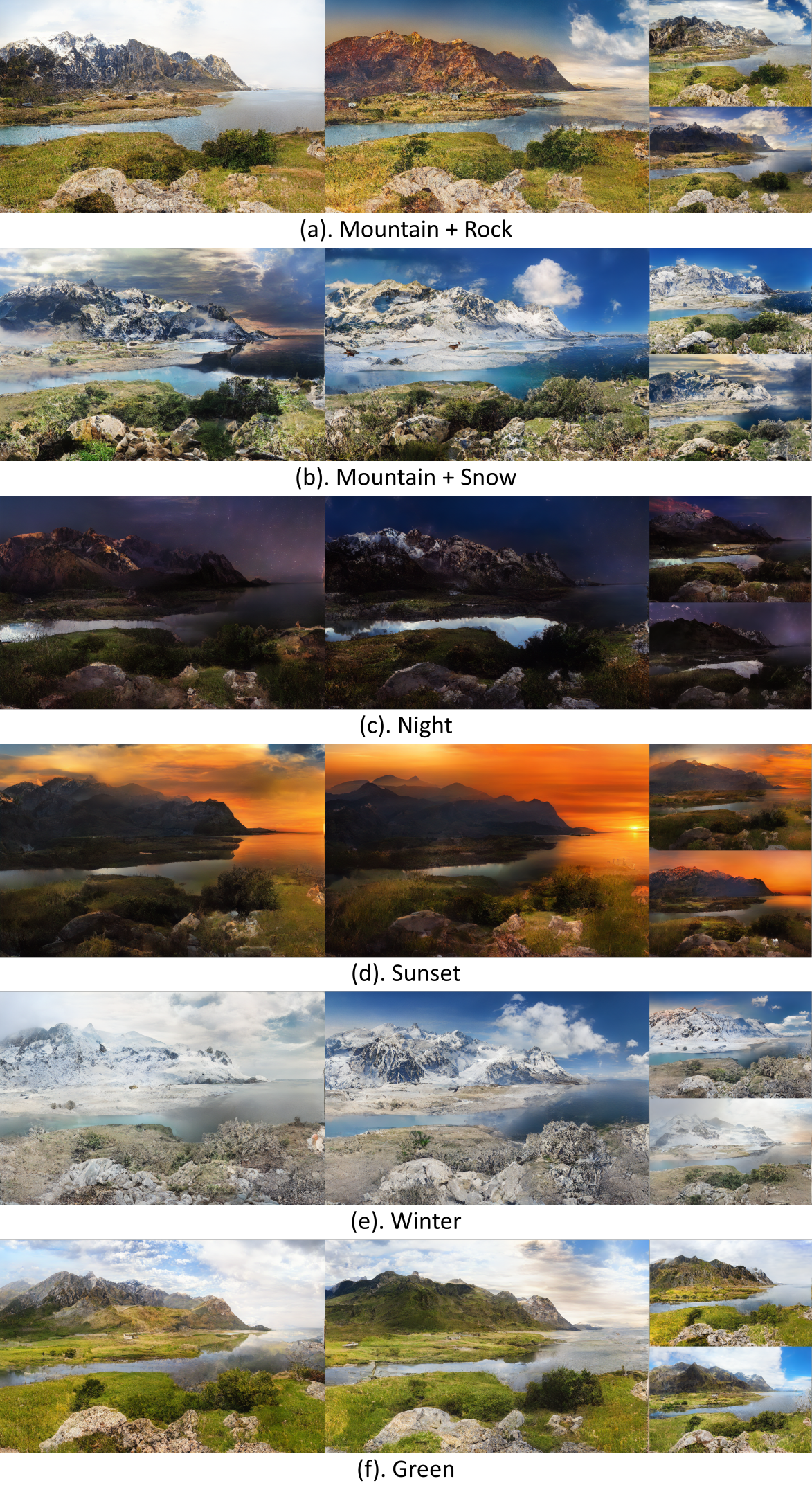}
\caption{{\bf Qualitative results for categorized styles.} We create six categories (a-f) and collect around 30 to 50 images each by keyword-search and manual inspection. Those images are used as style images to get average probabilities for categorized styles. Then we generate images of each category four times}
\label{fig:fig4}
\end{figure}

In the following experiment shown in Figure~\ref{fig:fig4}, we perform experiment on categorized set of style images. At first, we search images by keywords - ``mountain'' + ``rock'', ``mountain'' + ``snow'', ``night'', ``sunset'', and ``winter'' - in Pixabay website. Then we choose 30 to 50 images for each category. Additionally, category ``green'' is collected by manually selecting images of green fields. 
After collecting images by style category, we get the likelihood probabilities for every semantic labels from semantic regions in all style images.
Therefore, results in Figure~\ref{fig:fig4} well reflects the common styles of collected images in the category. In this figure, we repeat the inference four times for each category and contain all of them.

\begin{figure*}
\centering
\includegraphics[width=0.925\linewidth]{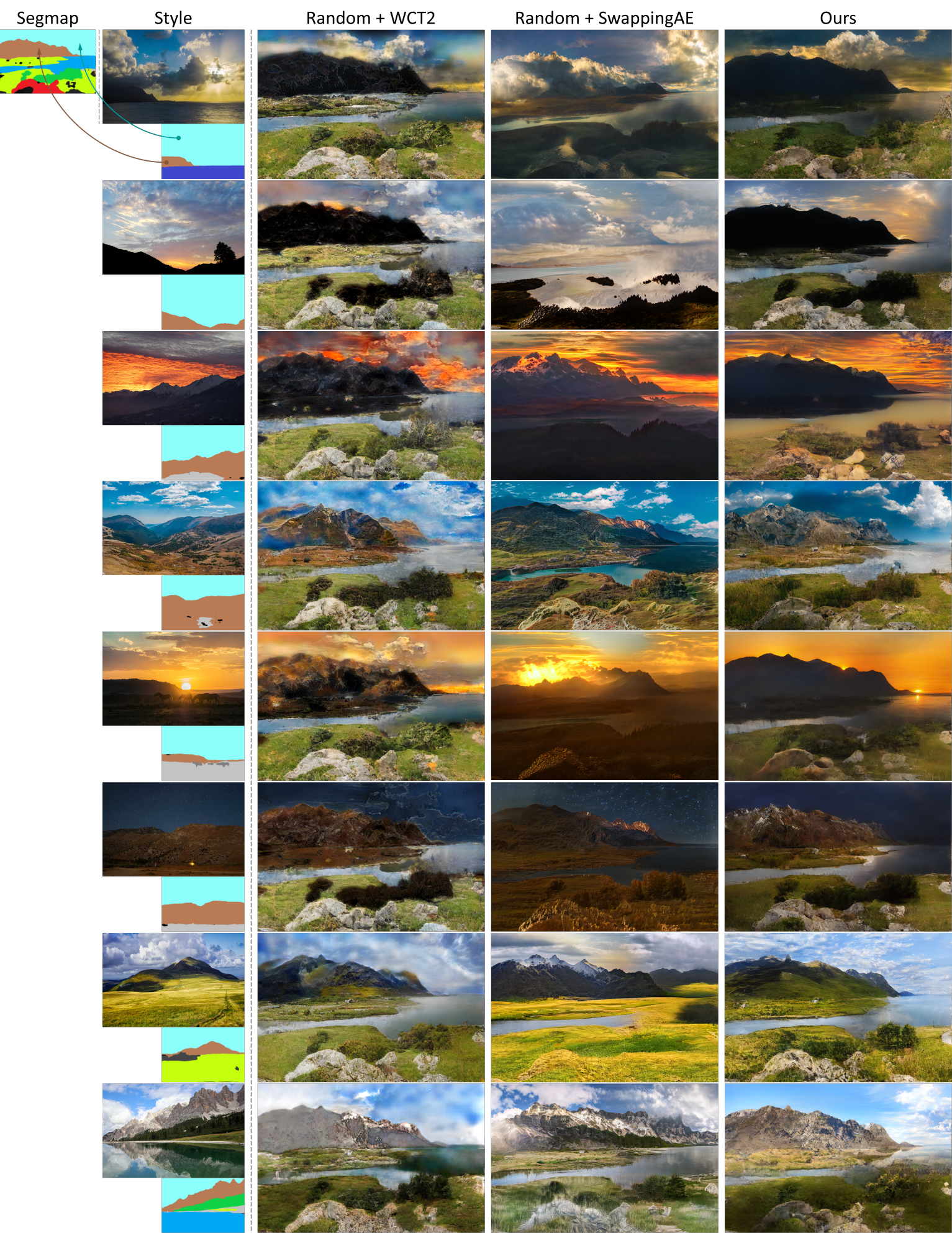}
\caption{{\bf Qualitative comparison with arbitrary style transfer.} Given a style images, our semantic-aware conditional generation shows plausible outputs in every semantic regions. ``Segmap'' is a semantic segmentation map to feed as a conditional input to baseline and our method for semantic-aware stylization. For comparison, we run semantic-aware style transfer, WCT2~\cite{yoo2019photorealistic}, and texture swapping, SwappingAE~\cite{park2020swapping} (official ``Flickr Mountains'' pretrained model), on randomly generated samples from the baseline method}
\label{fig:fig5}
\end{figure*}

To the best of our knowledge, this is the first conditional image generation using pre-trained generative model without learning such condition. To achieve a similar result using any existing method, we have a few workarounds. Firstly, there are a few researches on few-shot learning of image generative model, however, they requires hundreds of samples and hardly generate plausible samples on complex scenes. Instead, we can stylize the randomly generated samples using an arbitrary style transfer method. 
Since our method can apply style conditions by semantic regions, we compare it with one of the state-of-the-art arbitrary style transfer, WCT2~\cite{yoo2019photorealistic}, especially which can perform a semantic-aware style transfer. Also we prepare recently well-known texture transferring method, SwappingAE~\cite{park2020swapping}.
Although the style transfer method produces reasonable outputs in some cases, it does not consistently generate pleasant outputs in most cases. 
Furthermore, style transfer approach still yields unmatched or ruined color and texture, which is not desirable in high-resolution realistic scene generation.
SwappingAE is excellent at transferring tones and texture details. However, it fails to maintain semantic contents. Moreover, it tends to follow the spatial scene compositions of style image. (For example, it alters mountain to sky in the first two images, river to ground in the second last, and ground to river in the last image.)
Detailed results are shown in Figure~\ref{fig:fig5}.

\begin{figure}
\centering
\includegraphics[width=0.9\linewidth]{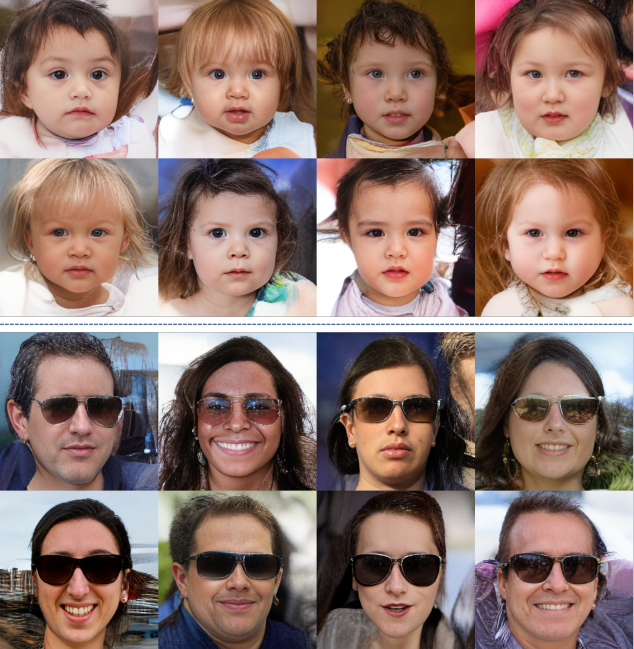}
\caption{{\bf FFHQ babies/sunglasses-guided samples using our method.} (Top): Babies-guided random samples. (Bottom): Sunglasses-guided random samples.}
\label{fig:fig6}
\end{figure}

\begin{table}
  \begin{center}
    {\small{
\begin{tabular}{lll}
\toprule
Methods & Babies & Sunglasses \\
\midrule
{\bf Random sampling } & 155.30 & 94.21 \\
{\bf \cite{ojha2021few} } & 74.39 & 42.13 \\
{\bf Ours } & {\bf 49.96 } & {\bf 39.15 } \\
\bottomrule
\end{tabular}
}}
\end{center}
\caption{{\bf Quantitative evaluation on FFHQ.} FID scores on two categories of FFHQ dataset are measured. Our work synthesizes images that are closer to the category of given guidance than the state-of-the-art competitor~\cite{ojha2021few}. ``Random sampling'' means random sampling using unconditional baseline~\cite{esser2021taming}.}
\label{table:quantitative_results}
\end{table}

Our work uses pseudo-segmentation map which is already available for semantic image synthesis of our baseline work~\cite{esser2021taming}. Having a semantic segmentation map improves extracting the exact per-semantic style and lets us partially stylize the output. On the contrary, in cases of unconditional face synthesis, we can use aligned dataset and collect the probability vectors by spatial locations.
To perform a quantitative assessment of an unconditional model, we measure Frechet Inception Distance (FID)~\cite{heusel2017gans} score on Flickr-Faces-HQ (FFHQ)~\cite{karras2019style} dataset and compare with the state-of-the-art few shot image generation~\cite{ojha2021few}. To use as a guidance, we randomly select 10 images of baby and sunglasses each. Then we calculate FID using 2492 and 2683 images of baby and sunglasses each which are the same test sets made public by~\cite{ojha2021few}. We use the same face synthesis model that the baseline work~\cite{esser2021taming} released.
Quantitative results of guided-inference of unconditional transformer are given in Table~\ref{table:quantitative_results}. 
One limitation is, however, our proposed method is not able to stylize out of learnt domain because our method guides the inference of transformer only, not re-training the model. Thus, in this experiment, we exclude ``Sketches'' category from the experiment shown in the competitor~\cite{ojha2021few}. 

%===========================================================
\section{Limitations}

Our method provides a style image to guide the inference of transformer introduced in our baseline work~\cite{esser2021taming}. To achieve similar effect, most of the recent studies require a re-training process and thousands of style images as well, and this makes recent studies harder to apply to the cases where only one or dozens of style images are provided. Our paper is unique in that, instead of requiring a re-training process, it only needs a single or a few style images to get the output of the desired style, while it inherits advantages of the pre-trained backbone. However, since it does not include a re-training process, it cannot generate samples out of the distribution of the training dataset. Therefore, as shown in Table~\ref{table:quantitative_results}, it is impossible to generate ``Sketches'' style output using the model trained on real face images.

\begin{table}
  \begin{center}
    {\small{
\begin{tabular}{lcc}
\toprule
Styles & 
\begin{tabular}{@{}c@{}} {\bf Random sampling } \end{tabular} &
\begin{tabular}{@{}c@{}} {\bf Ours } \end{tabular} \\
\midrule
Abstract Expressionism & 94.24 & {\bf 80.82 } \\
Art Nouveau Modern & 65.73 & {\bf 58.13 } \\
Baroque & 83.02 & {\bf 68.63 } \\
Cubism & 96.18 & {\bf 69.45 } \\
Naive Art Primitivism & 79.25 & {\bf 70.58 } \\
Northern Renaissance & 84.42 & {\bf 80.21 }\\
Rococo & 96.79 & {\bf 82.26 } \\
\bottomrule
\end{tabular}
}}
\end{center}
\caption{{\bf Ablation study of unconditional image synthesis on WikiArt dataset~\cite{Wikiartdataset}.} Unlike above experiments, in this experiment, our method is not provided any spatial aid for style likelihood. FID scores on the styles of the art by the history are measured. We only count styles that have more than 2000 images and less than 4500 images to prevent biased random sampling. ``Baseline'' stands for random sampling-based synthesis of baseline transformer, while ``Ours'' does for our style-guided synthesis.}
\label{table:quantitative_results2}
\end{table}

To model the ill-posed $P(u|T)$ with the known variables, we use probability distributions of $P(z_u)$ and $P_D(T)$. To improve accuracy in the process, we use semantic label maps in case of landscape synthesis and coordinates of a latent map in case of aligned face synthesis. 
However, if we aggregate probability distribution over all entries of a latent map without spatial separation, it generates a limited result. To show this limitation, where we claim the minimal performance, we carry out an ablation study on WikiArt dataset~\cite{Wikiartdataset}. In this experiment of unconditional image synthesis, we collect style vectors from the entire region of the image. 
As shown in Table~\ref{table:quantitative_results2}, FID scores using our method are lower than random sampling-based synthesis. However, it is not dramatic gab like proven in Table~\ref{table:quantitative_results}.

%===========================================================
\section{Conclusion and Future Work}

We believe that this is the first study on the guided inference of transformer to accept an additional arbitrary condition to stylize the output without requiring any parameter nor training procedure. To this end, we interpret probability distribution conditioned on style to a product of style distribution and a prior from transformer. 
In the experiment section, we prove our method on landscape dataset where styles are diverse and distinguishable. Compared to random generation of the original architecture~\cite{esser2021taming}, our method can generate scenes of an arbitrary style that are rare from the original method such as night scene. As an additional arbitrary condition, our work can use a single style image as well as a categorized set of style images. 
Our work is also similar to style transfer in a way that it stylizes the output image. However, our method uses the pretrained architecture and lets it describe target condition without re-training nor requiring any additional architecture, and thus our method does not need a condition-labeled dataset either. 

Since our approach is built upon autoregressive latent code prediction, our work is currently applied on a limited previous work. In the future, we therefore plan to build a general conditional autoregressive module to support an arbitrary conditional input. In addition, our work relies on transformer to build the proper sequence, not the relative layout or sequence of latent codes of style image. Therefore, for more accurate spatial application of style, we will work on space-aware style-guided inference of transformer in the near future.

{\small
\bibliographystyle{ieee_fullname}
\bibliography{SGIT}

\begin{thebibliography}{10}\itemsep=-1pt

\bibitem{arik2017deep}
Sercan~{\"O} Ar{\i}k, Mike Chrzanowski, Adam Coates, Gregory Diamos, Andrew
  Gibiansky, Yongguo Kang, Xian Li, John Miller, Andrew Ng, Jonathan Raiman,
  et~al.
\newblock Deep voice: Real-time neural text-to-speech.
\newblock In {\em International Conference on Machine Learning}, pages
  195--204. PMLR, 2017.

\bibitem{arjovsky2017wasserstein}
Martin Arjovsky, Soumith Chintala, and L{\'e}on Bottou.
\newblock Wasserstein generative adversarial networks.
\newblock In {\em International Conference on Machine Learning}, pages
  214--223. PMLR, 2017.

\bibitem{berthelot2017began}
David Berthelot, Thomas Schumm, and Luke Metz.
\newblock Began: Boundary equilibrium generative adversarial networks.
\newblock {\em arXiv preprint arXiv:1703.10717}, 2017.

\bibitem{brock2018large}
Andrew Brock, Jeff Donahue, and Karen Simonyan.
\newblock Large scale gan training for high fidelity natural image synthesis.
\newblock In {\em International Conference on Learning Representations}, 2018.

\bibitem{chen2017rethinking}
Liang-Chieh Chen, George Papandreou, Florian Schroff, and Hartwig Adam.
\newblock Rethinking atrous convolution for semantic image segmentation.
\newblock {\em arXiv preprint arXiv:1706.05587}, 2017.

\bibitem{chen2018pixelsnail}
Xi Chen, Nikhil Mishra, Mostafa Rohaninejad, and Pieter Abbeel.
\newblock Pixelsnail: An improved autoregressive generative model.
\newblock In {\em International Conference on Machine Learning}, pages
  864--872. PMLR, 2018.

\bibitem{cheng2021style}
Jiaxin Cheng, Ayush Jaiswal, Yue Wu, Pradeep Natarajan, and Prem Natarajan.
\newblock Style-aware normalized loss for improving arbitrary style transfer.
\newblock In {\em Proceedings of the IEEE/CVF Conference on Computer Vision and
  Pattern Recognition}, pages 134--143, 2021.

\bibitem{Wikiartdataset}
Cs-Chan.
\newblock Artgan/wikiart dataset.
\newblock https://github.com/cs-chan/ArtGAN/tree/master/WikiArt Dataset.

\bibitem{dundar2020panoptic}
Aysegul Dundar, Karan Sapra, Guilin Liu, Andrew Tao, and Bryan Catanzaro.
\newblock Panoptic-based image synthesis.
\newblock In {\em Proceedings of the IEEE/CVF Conference on Computer Vision and
  Pattern Recognition}, pages 8070--8079, 2020.

\bibitem{esser2021taming}
Patrick Esser, Robin Rombach, and Bjorn Ommer.
\newblock Taming transformers for high-resolution image synthesis.
\newblock In {\em Proceedings of the IEEE/CVF Conference on Computer Vision and
  Pattern Recognition}, pages 12873--12883, 2021.

\bibitem{goodfellow2014generative}
Ian Goodfellow, Jean Pouget-Abadie, Mehdi Mirza, Bing Xu, David Warde-Farley,
  Sherjil Ozair, Aaron Courville, and Yoshua Bengio.
\newblock Generative adversarial nets.
\newblock {\em Advances in Neural Information Processing Systems}, 27, 2014.

\bibitem{heusel2017gans}
Martin Heusel, Hubert Ramsauer, Thomas Unterthiner, Bernhard Nessler, and Sepp
  Hochreiter.
\newblock Gans trained by a two time-scale update rule converge to a local nash
  equilibrium.
\newblock {\em Advances in Neural Information Processing Systems}, 30, 2017.

\bibitem{hong2018inferring}
Seunghoon Hong, Dingdong Yang, Jongwook Choi, and Honglak Lee.
\newblock Inferring semantic layout for hierarchical text-to-image synthesis.
\newblock In {\em Proceedings of the IEEE Conference on Computer Vision and
  Pattern Recognition}, pages 7986--7994, 2018.

\bibitem{hudson2021ganformer}
Drew~A Hudson and C.~Lawrence Zitnick.
\newblock Generative adversarial transformers.
\newblock In {\em International Conference on Machine Learning}, pages
  4487--4499. PMLR, 2021.

\bibitem{karras2018progressive}
Tero Karras, Timo Aila, Samuli Laine, and Jaakko Lehtinen.
\newblock Progressive growing of gans for improved quality, stability, and
  variation.
\newblock In {\em International Conference on Learning Representations}, 2018.

\bibitem{karras2019style}
Tero Karras, Samuli Laine, and Timo Aila.
\newblock A style-based generator architecture for generative adversarial
  networks.
\newblock In {\em Proceedings of the IEEE/CVF Conference on Computer Vision and
  Pattern Recognition}, pages 4401--4410, 2019.

\bibitem{kingma2013auto}
Diederik~P Kingma and Max Welling.
\newblock Auto-encoding variational bayes.
\newblock {\em International Conference on Learning Representations}, 2014.

\bibitem{li2017universal}
Yijun Li, Chen Fang, Jimei Yang, Zhaowen Wang, Xin Lu, and Ming-Hsuan Yang.
\newblock Universal style transfer via feature transforms.
\newblock {\em Advances in Neural Information Processing Systems}, 30, 2017.

\bibitem{li2018closed}
Yijun Li, Ming-Yu Liu, Xueting Li, Ming-Hsuan Yang, and Jan Kautz.
\newblock A closed-form solution to photorealistic image stylization.
\newblock In {\em Proceedings of the European Conference on Computer Vision
  (ECCV)}, pages 453--468, 2018.

\bibitem{liu2019learning}
Xihui Liu, Guojun Yin, Jing Shao, Xiaogang Wang, and Hongsheng Li.
\newblock Learning to predict layout-to-image conditional convolutions for
  semantic image synthesis.
\newblock In {\em Proceedings of the 33rd International Conference on Neural
  Information Processing Systems}, pages 570--580, 2019.

\bibitem{luan2017deep}
Fujun Luan, Sylvain Paris, Eli Shechtman, and Kavita Bala.
\newblock Deep photo style transfer.
\newblock In {\em Proceedings of the IEEE Conference on Computer Vision and
  Pattern Recognition}, pages 4990--4998, 2017.

\bibitem{mao2017least}
Xudong Mao, Qing Li, Haoran Xie, Raymond~YK Lau, Zhen Wang, and Stephen
  Paul~Smolley.
\newblock Least squares generative adversarial networks.
\newblock In {\em Proceedings of the IEEE international conference on computer
  vision}, pages 2794--2802, 2017.

\bibitem{mescheder2018training}
Lars Mescheder, Andreas Geiger, and Sebastian Nowozin.
\newblock Which training methods for gans do actually converge?
\newblock In {\em International Conference on Machine Learning}, pages
  3481--3490. PMLR, 2018.

\bibitem{mirza2014conditional}
Mehdi Mirza and Simon Osindero.
\newblock Conditional generative adversarial nets.
\newblock {\em arXiv preprint arXiv:1411.1784}, 2014.

\bibitem{miyato2018spectral}
Takeru Miyato, Toshiki Kataoka, Masanori Koyama, and Yuichi Yoshida.
\newblock Spectral normalization for generative adversarial networks.
\newblock In {\em International Conference on Learning Representations}, 2018.

\bibitem{miyato2018cgans}
Takeru Miyato and Masanori Koyama.
\newblock cgans with projection discriminator.
\newblock In {\em International Conference on Learning Representations}, 2018.

\bibitem{odena2017conditional}
Augustus Odena, Christopher Olah, and Jonathon Shlens.
\newblock Conditional image synthesis with auxiliary classifier gans.
\newblock In {\em International Conference on Machine Learning}, pages
  2642--2651. PMLR, 2017.

\bibitem{ojha2021few}
Utkarsh Ojha, Yijun Li, Jingwan Lu, Alexei~A Efros, Yong~Jae Lee, Eli
  Shechtman, and Richard Zhang.
\newblock Few-shot image generation via cross-domain correspondence.
\newblock In {\em Proceedings of the IEEE/CVF Conference on Computer Vision and
  Pattern Recognition}, pages 10743--10752, 2021.

\bibitem{oord2018parallel}
Aaron Oord, Yazhe Li, Igor Babuschkin, Karen Simonyan, Oriol Vinyals, Koray
  Kavukcuoglu, George Driessche, Edward Lockhart, Luis Cobo, Florian Stimberg,
  et~al.
\newblock Parallel wavenet: Fast high-fidelity speech synthesis.
\newblock In {\em International Conference on Machine Learning}, pages
  3918--3926. PMLR, 2018.

\bibitem{oord2016wavenet}
Aaron van~den Oord, Sander Dieleman, Heiga Zen, Karen Simonyan, Oriol Vinyals,
  Alex Graves, Nal Kalchbrenner, Andrew Senior, and Koray Kavukcuoglu.
\newblock Wavenet: A generative model for raw audio.
\newblock In {\em 9th ISCA Speech Synthesis Workshop}, pages 125--125, 2016.

\bibitem{oord2016conditional}
A{\"a}ron van~den Oord, Nal Kalchbrenner, Oriol Vinyals, Lasse Espeholt, Alex
  Graves, and Koray Kavukcuoglu.
\newblock Conditional image generation with pixelcnn decoders.
\newblock In {\em Proceedings of the 30th International Conference on Neural
  Information Processing Systems}, pages 4797--4805, 2016.

\bibitem{park2019semantic}
Taesung Park, Ming-Yu Liu, Ting-Chun Wang, and Jun-Yan Zhu.
\newblock Semantic image synthesis with spatially-adaptive normalization.
\newblock In {\em Proceedings of the IEEE/CVF Conference on Computer Vision and
  Pattern Recognition}, pages 2337--2346, 2019.

\bibitem{park2020swapping}
Taesung Park, Jun-Yan Zhu, Oliver Wang, Jingwan Lu, Eli Shechtman, Alexei
  Efros, and Richard Zhang.
\newblock Swapping autoencoder for deep image manipulation.
\newblock {\em Advances in Neural Information Processing Systems},
  33:7198--7211, 2020.

\bibitem{parmar2018image}
Niki Parmar, Ashish Vaswani, Jakob Uszkoreit, Lukasz Kaiser, Noam Shazeer,
  Alexander Ku, and Dustin Tran.
\newblock Image transformer.
\newblock In {\em International Conference on Machine Learning}, pages
  4055--4064. PMLR, 2018.

\bibitem{radford2015unsupervised}
Alec Radford, Luke Metz, and Soumith Chintala.
\newblock Unsupervised representation learning with deep convolutional
  generative adversarial networks.
\newblock In {\em International Conference on Learning Representations}, 2016.

\bibitem{razavi2019generating}
Ali Razavi, Aaron van~den Oord, and Oriol Vinyals.
\newblock Generating diverse high-fidelity images with vq-vae-2.
\newblock In {\em Advances in Neural Information Processing Systems}, pages
  14866--14876, 2019.

\bibitem{reed2016generative}
Scott Reed, Zeynep Akata, Xinchen Yan, Lajanugen Logeswaran, Bernt Schiele, and
  Honglak Lee.
\newblock Generative adversarial text to image synthesis.
\newblock In {\em International Conference on Machine Learning}, pages
  1060--1069. PMLR, 2016.

\bibitem{salimans2016improved}
Tim Salimans, Ian Goodfellow, Wojciech Zaremba, Vicki Cheung, Alec Radford, and
  Xi Chen.
\newblock Improved techniques for training gans.
\newblock In {\em Proceedings of the 30th International Conference on Neural
  Information Processing Systems}, pages 2234--2242, 2016.

\bibitem{shen2018natural}
Jonathan Shen, Ruoming Pang, Ron~J Weiss, Mike Schuster, Navdeep Jaitly,
  Zongheng Yang, Zhifeng Chen, Yu Zhang, Yuxuan Wang, Rj Skerrv-Ryan, et~al.
\newblock Natural tts synthesis by conditioning wavenet on mel spectrogram
  predictions.
\newblock In {\em 2018 IEEE International Conference on Acoustics, Speech and
  Signal Processing (ICASSP)}, pages 4779--4783. IEEE, 2018.

\bibitem{sheng2018avatar}
Lu Sheng, Ziyi Lin, Jing Shao, and Xiaogang Wang.
\newblock Avatar-net: Multi-scale zero-shot style transfer by feature
  decoration.
\newblock In {\em Proceedings of the IEEE Conference on Computer Vision and
  Pattern Recognition}, pages 8242--8250, 2018.

\bibitem{tan2021diverse}
Zhentao Tan, Menglei Chai, Dongdong Chen, Jing Liao, Qi Chu, Bin Liu, Gang Hua,
  and Nenghai Yu.
\newblock Diverse semantic image synthesis via probability distribution
  modeling.
\newblock In {\em Proceedings of the IEEE/CVF Conference on Computer Vision and
  Pattern Recognition}, pages 7962--7971, 2021.

\bibitem{tan2021efficient}
Zhentao Tan, Dongdong Chen, Qi Chu, Menglei Chai, Jing Liao, Mingming He, Lu
  Yuan, Gang Hua, and Nenghai Yu.
\newblock Efficient semantic image synthesis via class-adaptive normalization.
\newblock {\em IEEE Transactions on Pattern Analysis and Machine Intelligence},
  2021.

\bibitem{tang2020dual}
Hao Tang, Song Bai, and Nicu Sebe.
\newblock Dual attention gans for semantic image synthesis.
\newblock In {\em Proceedings of the 28th ACM International Conference on
  Multimedia}, pages 1994--2002, 2020.

\bibitem{van2017neural}
Aaron van~den Oord, Oriol Vinyals, and Koray Kavukcuoglu.
\newblock Neural discrete representation learning.
\newblock In {\em Proceedings of the 31st International Conference on Neural
  Information Processing Systems}, pages 6309--6318, 2017.

\bibitem{van2016pixel}
Aaron Van~Oord, Nal Kalchbrenner, and Koray Kavukcuoglu.
\newblock Pixel recurrent neural networks.
\newblock In {\em International Conference on Machine Learning}, pages
  1747--1756. PMLR, 2016.

\bibitem{vaswani2017attention}
Ashish Vaswani, Noam Shazeer, Niki Parmar, Jakob Uszkoreit, Llion Jones,
  Aidan~N Gomez, {\L}ukasz Kaiser, and Illia Polosukhin.
\newblock Attention is all you need.
\newblock In {\em Advances in Neural Information Processing Systems}, pages
  5998--6008, 2017.

\bibitem{wang2018high}
Ting-Chun Wang, Ming-Yu Liu, Jun-Yan Zhu, Andrew Tao, Jan Kautz, and Bryan
  Catanzaro.
\newblock High-resolution image synthesis and semantic manipulation with
  conditional gans.
\newblock In {\em Proceedings of the IEEE conference on computer vision and
  pattern recognition}, pages 8798--8807, 2018.

\bibitem{wang2017tacotron}
Yuxuan Wang, R.~J. Skerry-Ryan, Daisy Stanton, Yonghui Wu, Ron~J. Weiss,
  Navdeep Jaitly, Zongheng Yang, Ying Xiao, Zhifeng Chen, Samy Bengio, Quoc~V.
  Le, Yannis Agiomyrgiannakis, Rob Clark, and Rif~A. Saurous.
\newblock Tacotron: Towards end-to-end speech synthesis.
\newblock In {\em Interspeech 2017}, pages 4006--4010, 2017.

\bibitem{xu2018attngan}
Tao Xu, Pengchuan Zhang, Qiuyuan Huang, Han Zhang, Zhe Gan, Xiaolei Huang, and
  Xiaodong He.
\newblock Attngan: Fine-grained text to image generation with attentional
  generative adversarial networks.
\newblock In {\em Proceedings of the IEEE conference on computer vision and
  pattern recognition}, pages 1316--1324, 2018.

\bibitem{yim2020filter}
Jonghwa Yim, Jisung Yoo, Won-joon Do, Beomsu Kim, and Jihwan Choe.
\newblock Filter style transfer between photos.
\newblock In {\em European Conference on Computer Vision}, pages 103--119.
  Springer, 2020.

\bibitem{yoo2019photorealistic}
Jaejun Yoo, Youngjung Uh, Sanghyuk Chun, Byeongkyu Kang, and Jung-Woo Ha.
\newblock Photorealistic style transfer via wavelet transforms.
\newblock In {\em Proceedings of the IEEE International Conference on Computer
  Vision}, pages 9036--9045, 2019.

\bibitem{zhang2019self}
Han Zhang, Ian Goodfellow, Dimitris Metaxas, and Augustus Odena.
\newblock Self-attention generative adversarial networks.
\newblock In {\em International Conference on Machine Learning}, pages
  7354--7363. PMLR, 2019.

\bibitem{zhang2022resnest}
Hang Zhang, Chongruo Wu, Zhongyue Zhang, Yi Zhu, Haibin Lin, Zhi Zhang, Yue
  Sun, Tong He, Jonas Mueller, R Manmatha, et~al.
\newblock Resnest: Split-attention networks.
\newblock In {\em Proceedings of the IEEE/CVF Conference on Computer Vision and
  Pattern Recognition}, pages 2736--2746, 2022.

\bibitem{zhang2017stackgan}
Han Zhang, Tao Xu, Hongsheng Li, Shaoting Zhang, Xiaogang Wang, Xiaolei Huang,
  and Dimitris~N Metaxas.
\newblock Stackgan: Text to photo-realistic image synthesis with stacked
  generative adversarial networks.
\newblock In {\em Proceedings of the IEEE international conference on computer
  vision}, pages 5907--5915, 2017.

\bibitem{zheng2020example}
Haitian Zheng, Haofu Liao, Lele Chen, Wei Xiong, Tianlang Chen, and Jiebo Luo.
\newblock Example-guided image synthesis using masked spatial-channel attention
  and self-supervision.
\newblock In {\em European Conference on Computer Vision}, pages 422--439.
  Springer, 2020.

\bibitem{zhu2020sean}
Peihao Zhu, Rameen Abdal, Yipeng Qin, and Peter Wonka.
\newblock Sean: Image synthesis with semantic region-adaptive normalization.
\newblock In {\em Proceedings of the IEEE/CVF Conference on Computer Vision and
  Pattern Recognition}, pages 5104--5113, 2020.

\bibitem{zhu2020semantically}
Zhen Zhu, Zhiliang Xu, Ansheng You, and Xiang Bai.
\newblock Semantically multi-modal image synthesis.
\newblock In {\em Proceedings of the IEEE/CVF Conference on Computer Vision and
  Pattern Recognition}, pages 5467--5476, 2020.

\end{thebibliography}
}

\end{document}